\documentclass[runningheads]{llncs}
\usepackage{graphicx}
\usepackage{amsmath,amssymb} 
\usepackage{color}
\usepackage[width=122mm,left=12mm,paperwidth=146mm,height=193mm,top=12mm,paperheight=217mm]{geometry}

\usepackage{microtype} 
\usepackage{tabularx}
\usepackage{booktabs}
\usepackage{rotating}
\usepackage{authblk}

\makeatletter
\newcommand{\@chapapp}{\relax}%
\makeatother
\usepackage[title,toc,titletoc,header]{appendix}

\usepackage{hyperref}

\newcommand{\deepcut}{\textit{DeepCut}}

\newcommand{\deepercut}{\textit{DeeperCut}}

\newcommand{\myparagraph}[1]{\vspace{0.1em}\noindent\textbf{#1}}

\newcommand{\myurl}{\url{http://pose.mpi-inf.mpg.de}}

\begin{document}
\pagestyle{headings}
\mainmatter

\title{DeeperCut: A Deeper, Stronger, and Faster Multi-Person Pose Estimation Model} 

\author{Eldar Insafutdinov$^1$, Leonid Pishchulin$^1$, Bjoern Andres$^1$, \\Mykhaylo Andriluka$^{1,2}$, and Bernt Schiele$^1$}
\institute{$^1$Max Planck Institute for Informatics, Germany,\\$^2$Stanford University, USA}
\titlerunning{A Deeper, Stronger, and Faster Multi-Person Pose Estimation Model}
\authorrunning{E. Insafutdinov, L. Pishchulin, B. Andres, M. Andriluka, and B. Schiele}


\maketitle

\begin{abstract}
  The goal of this paper is to advance the state-of-the-art of
  articulated pose estimation in scenes with multiple people. To that
  end we contribute on three fronts. We propose (1) improved body part
  detectors that generate effective bottom-up proposals for body
  parts; (2) novel image-conditioned pairwise terms that allow to
  assemble the proposals into a variable number of consistent body
  part configurations; and (3) an incremental optimization strategy
  that explores the search space more efficiently thus leading both to
  better performance and significant speed-up factors. Evaluation is
  done on two single-person and two multi-person pose estimation
  benchmarks. The proposed approach significantly outperforms best
  known multi-person pose estimation results while demonstrating
  competitive performance on the task of single person pose
  estimation\footnote{Models and code available at \myurl}.


\end{abstract}

\section{Introduction}
\label{sec:intro}

\begin{figure}[t]
	\centering
	\begin{tabular}{c c c}
	  \includegraphics[height=0.19\linewidth]{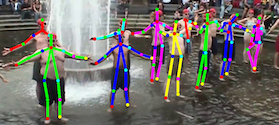}&
	  \includegraphics[height=0.19\linewidth]{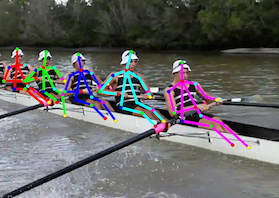}&
          \includegraphics[height=0.19\linewidth]{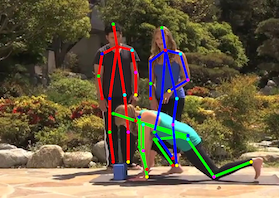} \\
        \end{tabular}
        \begin{tabular}{c c c}
     	  \includegraphics[height=0.2\linewidth]{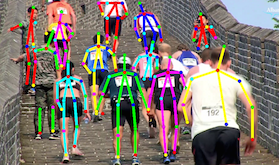} &
          \includegraphics[height=0.2\linewidth]{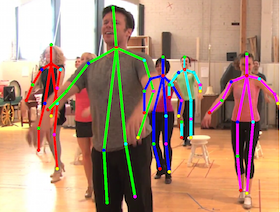} &
          \includegraphics[height=0.2\linewidth]{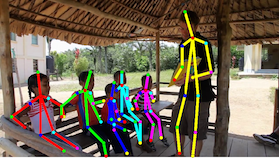}\\
	\end{tabular}
        \begin{tabular}{c c c}
          \includegraphics[height=0.203\linewidth]{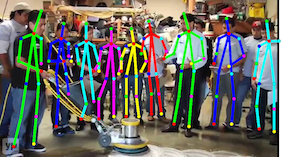} &
	  \includegraphics[height=0.203\linewidth]{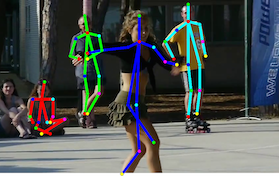} &
          \includegraphics[height=0.203\linewidth]{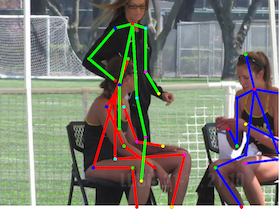} \\
	\end{tabular}
	\vspace{-1em}
	\caption{Sample multi-person pose estimation results by the
          proposed \deepercut.}
   \label{fig:teaser}
\end{figure}


Human pose estimation has recently made dramatic progress in
particular on standard benchmarks for single person pose
estimation~\cite{johnson10bmvc,andriluka14cvpr}. This progress has
been facilitated by the use of deep learning-based
architectures~\cite{krizhevsky12nips,Simonyan14c} and by the
availability of large-scale datasets such as ``MPII Human
Pose''~\cite{andriluka14cvpr}. In order to make further progress on
the challenging task of multi-person pose estimation we carefully
design and evaluate several key-ingredients for human pose estimation.

The first ingredient we consider is the generation of body part
hypotheses. Essentially all prominent pose estimation methods include
a component that detects body parts or estimates their
position. While early work used classifiers such as SVMs and
AdaBoost~\cite{johnson10bmvc,andriluka11ijcv,yang12pami,pishchulin13cvpr}, modern
approaches build on different flavors of deep learning-based
architectures~\cite{tompson14nips,chen14nips,pishchulin16cvpr,wei16cvpr}.
The second key ingredient are pairwise terms between body part
hypotheses that help grouping those into valid human pose configurations. 
In earlier models such pairwise terms were essential for
good performance~\cite{johnson10bmvc,andriluka11ijcv,yang12pami}.
Recent methods seem to profit less from such pairwise terms 
due to stronger unaries~\cite{tompson14nips,pishchulin16cvpr,wei16cvpr}.
Image-conditioned pairwise terms~\cite{pishchulin13cvpr,chen14nips}
however have the promise to allow for better grouping.
Last but not least, inference time is always a key consideration for
pose estimation models. Often, model complexity has to be treated for
speed and thus many models do not consider all spatial relations that
would be beneficial for best performance.

In this paper we contribute to all three aspects and thereby
significantly push the state of the art in multi-person pose
estimation. We use a general optimization framework introduced in our
previous work~\cite{pishchulin16cvpr} as a test bed for all three key
ingredients proposed in this paper, as it allows to easily replace and
combine different components. Our contributions are three-fold,
leading to a novel multi-person pose estimation approach that is
deeper, stronger, and faster compared to the state of the
art~\cite{pishchulin16cvpr}:
\begin{itemize}
\item ``deeper'': we propose strong body part detectors based on
  recent advances in deep learning \cite{he2015deep} that -- taken
  alone -- already allow to obtain competitive performance on pose
  estimation benchmarks.
\item ``stronger'': we introduce novel image-conditioned pairwise terms between body
  parts that allow to push performance in the challenging case of
  multi-people pose estimation.
\item ``faster'': we demonstrate that using our image-conditioned pairwise
  along with very good part detection candidates in a fully-connected
  model dramatically reduces the run-time by $2$--$3$ orders of
  magnitude. Finally, we introduce a novel incremental optimization
  method to achieve a further $4$x run-time reduction while improving
  human pose estimation accuracy.
\end{itemize}
We evaluate our approach on two single-person and two multi-person
pose estimation benchmarks and report the best results in each
case. Sample multi-person pose estimation predictions by the proposed
approach are shown in Fig.~\ref{fig:teaser}.

\myparagraph{Related work.} Articulated human pose estimation has been traditionally formulated as a
structured prediction task that requires an inference step combining local observations of body
joints with spatial constraints. Various formulations have been proposed based on tree
\cite{Ramanan:2006:LPI,Jiang:2008:GPE,johnson11cvpr,yang12pami} and non-tree models
\cite{tran10eccv,wang13cvpr}. The goal of the inference process has been to refine observations from
local part detectors into coherent estimates of body
configurations. Models of this type have been increasingly superseded
by strong body part
detectors \cite{pishchulin13iccv,Gkioxari:2013:APE,Tompson:2015:EOL},
which has been reinforced by the development of strong image
representations based on convolutional networks. Recent work aimed to
incorporate convolutional detectors into part-based
models \cite{chen14nips} or design stronger detectors by combining the
detector output with location-based features
\cite{ramakrishna14eccv}.

Specifically, as we suggest in \cite{pishchulin16cvpr}, in the presence of strong detectors spatial
reasoning results in diminishing returns because most contextual information can be incorporated
directly in the detector. In this work we elevate the task to a new level of complexity by
addressing images with multiple potentially overlapping people. This results in a more complex
structured prediction problem with a variable number of outputs. In this setting we observe a large
boost from conducting inference on top of state-of-the-art part detectors.

Combining spatial models with convnets allows to increase the receptive field that is used for
inferring body joint locations. For example \cite{wei16cvpr} iteratively trains a cascade of
convolutional parts detectors, each detector taking the scoremap of all parts from the previous stage.
This effectively increases the depth of the network and the receptive field is comparable to the 
entire person.
With the recent developments in object detection newer architectures are composed of a large number of layers and the receptive field is large automatically. In this paper, we introduce a detector based
on the recently proposed deep residual networks \cite{he2015deep}. This allows us to train a detector
with a large receptive field \cite{wei16cvpr} and to incorporate intermediate supervision.


The use of purely geometric pairwise terms is suboptimal as they do not take local image 
evidence into account and only penalize deviation from the expected joint location. Due to the inherent
articulation of body parts the expected location can only approximately guide the inference. 
While this can be sufficient when people are relatively distant from each other, for closely positioned people
more discriminative pairwise costs are essential. 
Two prior works \cite{pishchulin13cvpr,chen14nips} have introduced image-dependent pairwise terms between connected body parts. While \cite{pishchulin13cvpr} uses an intermediate representation
based on poselets our pairwise terms are conditioned directly on the image. 
\cite{chen14nips} clusters relative positions of adjacent joints into $T=11$ clusters, and assigns different labels to the part depending on which cluster it falls to. Subsequently a CNN is trained to predict this extended set of classes and later an SVM is used to select the maximum scoring joint pair relation.


Single person pose estimation has advanced considerably, but the setting is simplified. Here we
focus on the more challenging problem of multi-person pose estimation. Previous work has 
addressed this problem as sequence of person detection and pose estimation
\cite{eichner10eccv,Ladicky:2013:HPE,Chen:2015:POC}. \cite{eichner10eccv} use a detector for
initialization and reasoning across people, but rely on simple
geometric body part relationships and only reason about person-person
occlusions. \cite{Chen:2015:POC} focus on single partially occluded
people, and handle multi-person scenes akin
to \cite{yang12pami}. In \cite{pishchulin16cvpr} we propose to jointly
detect and estimate configurations, but rely on simple pairwise terms
only, which limits the performance and, as we show, results in
prohibitive inference time to fully explore the search space. Here, we
innovate on multiple fronts both in terms of speed and accuracy.

\section{DeepCut Recap}
\label{sec:deepcut}

This section summarizes $\deepcut$~\cite{pishchulin16cvpr} and how unary
and pairwise terms are used in this approach. $\deepcut$ is a
state-of-the-art approach to multi-person pose estimation based on integer
linear programming (ILP) that jointly estimates
poses of all people present in an image by minimizing a joint
objective. This objective aims to jointly partition and label an
initial pool of body part candidates into consistent sets of body-part
configurations corresponding to distinct people.
We use $\deepcut$ as a general optimization framework that allows 
to easily replace and combine different components. 

Specifically, $\deepcut$ starts from a set $D$ of \emph{body part
  candidates}, i.e.~putative detections of body parts in a given
image, and a set $C$ of \emph{body part classes}, e.g., head,
shoulder, knee. 
The set $D$ of part candidates is typically generated by body part detectors
and each candidate 
%
$d \in D$ has a {\it unary score} for every body part class $c \in C$. 
Based on these unary scores $\deepcut$ associates a 
cost or reward $\alpha_{dc} \in \mathbb{R}$ to be paid by all feasible solutions of the pose estimation problem for which the body part candidate $d$ is a body part of class $c$.

Additionally, for every pair of distinct body part candidates $d, d' \in D$ and every two body part classes $c, c' \in C$,
the {\it pairwise term} is used to generate a cost or reward $\beta_{dd'cc'} \in \mathbb{R}$ to be paid by all feasible solutions of the pose estimation problem for which the body part $d$, classified as $c$, and the body part $d'$, classified as $c'$, belong to the same person.

With respect to these sets and costs, the pose estimation problem is cast as an ILP in two classes of 01-variables:
Variables $x: D \times C \to \{0,1\}$ indicate by $x_{dc} = 1$ that body part candidate $d$ is of body part class $c$.
If, for a $d \in D$ and all $c \in C$, $x_{dc} = 0$, the body part candidate $d$ is suppressed.
Variables $y: \tbinom{D}{2} \to \{0,1\}$ indicate by $y_{dd'} = 1$ that body part candidates $d$ and $d'$ belong to the same person.
Additional variables and constraints described in
\cite{pishchulin16cvpr}
link the variables $x$ and $y$ to the costs and ensure that feasible solutions $(x, y)$ well-define 
a selection and classification of body part candidates as body part classes
as well as a clustering of body part candidates into distinct people.

The $\deepcut$ ILP is hard and hard to approximate, as it generalizes
the minimum cost multicut or correlation clustering problem which is
APX-hard \cite{bansal-2004,demaine-2006}.  Using the branch-and-cut
algorithm~\cite{pishchulin16cvpr} to compute
constant-factor approximative feasible solutions of instances of the
$\deepcut$ ILP is not necessarily practical.
In Sec.~\ref{sec:incremental} we propose an incremental optimization
approach that uses branch-and-cut algorithm to incrementally solve
several instances of ILP, which results into $4$--$5$x run-time
reduction with increased pose estimation accuracy.

\section{Part Detectors}
\label{sec:unary-pairwise}

As argued before, strong part detectors are an essential ingredient of
modern pose estimation methods. We propose and evaluate a deep
fully-convolutional human body part detection model drawing on
powerful recent ideas from semantic segmentation, object
classification~\cite{long2014fully,chen14semantic,he2015deep} and
human pose
estimation~\cite{Tompson:2015:EOL,pishchulin16cvpr,wei16cvpr}.

\subsection{Model}


\myparagraph{Architecture.} We build on the recent advances in object
classification and adapt the extremely deep Residual Network (ResNet)
~\cite{he2015deep} for human body part detection. This model achieved
excellent results on the recent ImageNet Object Classification
Challenge and specifically tackles the problem of vanishing gradients
by passing the state through identity layers and modeling residual
functions. Our best performing body part detection model has $152$
layers (c.f. Sec.~\ref{sec:unary-eval}) which is in line with the
findings of~\cite{he2015deep}.


\myparagraph{Stride.} Adapting ResNet for the sliding window-based
body part detection is not straight forward: converting ResNet to the
fully convolutional mode leads to a $32$ px stride which is too coarse
for precise part localization. In \cite{pishchulin16cvpr} we show that
using a stride of $8$ px leads to good part detection
results. Typically, spatial resolution can be recovered by either
introducing up-sampling \textit{deconvolutional} layers
\cite{long2014fully}, or blowing up the convolutional filters using
the \textit{hole algorithm}~\cite{chen14semantic}. The latter has
shown to perform better on the task of semantic segmentation.
However, using the \textit{hole algorithm} to recover the spatial
resolution of ResNet is infeasible due to memory constraints. For
instance, the $22$ residual blocks in the conv4 bank of ResNet-101
constitute the major part of the network and running it at stride $8$ px
does not fit the net into GPU memory\footnote{We use NVIDIA
  Tesla K40 GPU with 12 GB RAM}. We thus employ a hybrid
approach. First, we remove the final classification as well as
average pooling layer. Then, we decrease the stride of the first
convolutional layers of the conv5 bank from $2$ px to $1$ px to prevent
down-sampling. Next, we add holes to all $3$x$3$ convolutions in 
conv5 to preserve their receptive field. This reduces the stride of
the full CNN to $16$ px. Finally, we add deconvolutional layers for
$2$x up-sampling and connect the final output to the output of the
conv3 bank.

\myparagraph{Receptive field size.} A large receptive field size allows
to incorporate context when predicting locations of individual
body parts. \cite{tompson14nips,wei16cvpr} argue about the importance
of large receptive fields and propose a complex hierarchical
architecture predicting parts at multiple resolution levels. The extreme
depth of ResNet allows for a very large receptive field (on the
order of 1000 px compared to VGG's 400 px~\cite{Simonyan14c})
without the need of introducing complex hierarchical architectures. We
empirically find that re-scaling the original image such that an
upright standing person is $340$ px high leads to best performance.

\myparagraph{Intermediate supervision.} Providing additional
supervision addresses the problem of vanishing gradients in deep
neural networks~\cite{szegedy15cvpr,lee15aistats,wei16cvpr}. In
addition to that, \cite{wei16cvpr} reports that using part scoremaps
produced at intermediate stages as inputs for subsequent stages helps
to encode spatial relations between parts, while~\cite{Pfister15a} use
spatial fusion layers that learn an implicit spatial model. ResNets
address the first problem by introducing identity connections and
learning residual functions. To address the second concern, we make a
slightly different choice: we add part loss layers inside the conv4
bank of ResNet. We argue that it is not strictly necessary to use
scoremaps as inputs for the subsequent stages. The activations from
such intermediate predictions are different only up to a linear
transformation and contain all information about part presence that is
available at that stage of the network. In Sec.~\ref{sec:unary-eval}
we empirically show a consistent improvement of part detection
performance when including intermediate supervision.

\myparagraph{Loss functions.} We use sigmoid activations and cross
entropy loss function during training~\cite{pishchulin16cvpr}.
We perform location refinement by predicting offsets from the
locations on the scoremap grid to the ground truth joint locations~\cite{pishchulin16cvpr}.


\myparagraph{Training.} We use the publicly available ResNet
implementation (Caffe) and initialize from the ImageNet-pre-trained
models. We train networks with SGD for 1M iterations, starting with
the learning rate lr=$0.001$ for $10$k, then lr=$0.002$ for $420$k,
lr=$0.0002$ for $300$k and lr=$0.0001$ for 300k. This corresponds to
roughly $17$ epochs of the MPII~\cite{andriluka14cvpr} train
set. Finetuning from ImageNet takes two days on a \textit{single}
GPU. Batch normalization~\cite{ioffe2015batch} worsens performance, as
the batch size of 1 in fully convolutional training is not enough to
provide a reliable estimate of activation statistics. During training
we switch off collection of statistics and use the mean and variance
that were gathered on the ImageNet dataset.


\subsection{Evaluation of Part Detectors}
\label{sec:unary-eval}
\myparagraph{Datasets.}  We use three public datasets: ``Leeds Sports
Poses'' (LSP)~\cite{johnson10bmvc} (person-centric (PC) annotations);
``LSP Extended'' (LSPET) ~\cite{johnson11cvpr};
``MPII Human Pose'' (``Single
Person'')~\cite{andriluka14cvpr} consisting of $19185$ training and
$7247$ testing poses.
To evaluate on LSP we train part detectors on the union of MPII, LSPET
and LSP training sets. To evaluate on MPII Single Person we train on
MPII \emph{only}.

\myparagraph{Evaluation measures.} We use the standard ``Percentage of
Correct Keypoints (PCK)'' evaluation
metric~\cite{sapp13cvpr,Toshev:2014:DHP,tompson14nips} and evaluation
scripts from the web page of~\cite{andriluka14cvpr}.
In addition to PCK at fixed threshold, we report ``Area under Curve''
(AUC) computed for the entire range of PCK thresholds.

\myparagraph{Results on LSP.} The results are shown in
Tab.~\ref{tab:multicut:lsp}. ResNet-50 with $8$ px stride achieves
$87.8$\% PCK and $63.7$\% AUC. Increasing the stride size to $16$ px
and up-sampling the scoremaps by $2$x to compensate for the loss on
resolution slightly drops the performance to $87.2$\% PCK. This is
expected as up-sampling cannot fully compensate for the information
loss due to a larger stride. Larger stride minimizes memory
requirements, which allows for training a deeper ResNet-$152$. The
latter significantly increases the performance ($89.1$ vs. $87.2$\%
PCK, $65.1$ vs. $63.1$\% AUC), as it has larger model
capacity. Introducing intermediate supervision further improves the
performance to $90.1$\% PCK and $66.1$\% AUC, as it constraints the
network to learn useful representations in the early stages and uses
them in later stages for spatial disambiguation of parts.

The results are compared to the state of the art in
Tab.~\ref{tab:multicut:lsp}. Our best model significantly outperforms
\deepcut~\cite{pishchulin16cvpr} ($90.1$\% PCK vs. $87.1$\% PCK), as it relies 
on deeper detection architectures. Our model performs on par with the recent approach of Wei
et al.~\cite{wei16cvpr} ($90.1$ vs. 90.5\% PCK, $66.1$ vs. $65.4$
AUC). This is interesting, as they use a much more complex multi-scale
multi-stage architecture.

\tabcolsep 1.5pt
\begin{table}[tbp]
 \scriptsize
  \centering
  \begin{tabular}{@{} l c ccc ccc c|c@{}}
    \toprule
    Setting& Head   & Sho  & Elb & Wri & Hip & Knee & Ank & PCK & AUC\\       
    \midrule
    ResNet-50 (8 px)                & 96.9  & 90.3  & 85.0  & 81.5  & 88.6  & 87.3 & 84.8 & 87.8 & 63.7 \\ 
    ResNet-50 (16 px + 2x up-sample) & 96.7  & 89.8  & 84.6  & 80.4  & 89.3  & 86.4 & 82.8 & 87.2 & 63.1 \\ 
    ResNet-101 (16 px + 2x up-sample) & 96.9  & 91.2  & 85.8  & 82.6  & 90.9  & \textbf{90.2} & 85.9 & 89.1 & 64.6 \\ 
    ResNet-152 (16 px + 2x up-sample) & 97.4  & 91.7  & 85.7  & 82.4  & 90.1  & 89.2 & 86.9 & 89.1 & 65.1\\
    \quad + intermediate supervision  & 97.4  & \textbf{92.7}  & \textbf{87.5}  & \textbf{84.4}  & \textbf{91.5}  & 89.9 & 87.2 & 90.1 & \textbf{66.1} \\
    \midrule
    \deepcut~\cite{pishchulin16cvpr} & 97.0  & 91.0  & 83.8 & 78.1  & 91.0  & 86.7 & 82.0 & 87.1 & 63.5 \\
    Wei et al.~\cite{wei16cvpr} & \textbf{97.8}  & 92.5  & 87.0  & 83.9  & \textbf{91.5}  & 90.8 & \textbf{89.9} & \textbf{90.5} & 65.4 \\
    Tompson et al.~\cite{tompson14nips}& 90.6  & 79.2  & 67.9  & 63.4  & 69.5  & 71.0 & 64.2 & 72.3 & 47.3 \\
    Chen\&Yuille~\cite{chen14nips}& 91.8  & 78.2  & 71.8  & 65.5  & 73.3  & 70.2 & 63.4& 73.4 & 40.1 \\
    Fan et al.~\cite{fan15cvpr} & 92.4 & 75.2& 65.3& 64.0& 75.7& 68.3& 70.4& 73.0 & 43.2 \\
    \bottomrule
  \end{tabular}
  \caption[]{Pose estimation results (PCK) on LSP (PC) dataset.}
  \label{tab:multicut:lsp}
\vspace{-0.5cm}
\end{table}

\tabcolsep 1.5pt
\begin{table}[tbp]
 \scriptsize
  \centering
  \begin{tabular}{@{} l c ccc ccc c|c@{}}
    \toprule
    Setting& Head   & Sho  & Elb & Wri & Hip & Knee & Ank & PCK$_h$ & AUC\\
    \midrule
    ResNet-152                       & 96.3  & 94.1  & 88.6  & 83.9  & 87.2  & 82.9 & 77.8 & 87.8 & 60.0 \\
    \quad + intermediate supervision & 96.8  & \textbf{95.2} & \textbf{89.3}  & \textbf{84.4}  & \textbf{88.4}  & \textbf{83.4} & 78.0 & \textbf{88.5} & 60.8 \\
    \midrule
    \deepcut~\cite{pishchulin16cvpr} & 94.1 & 90.2  & 83.4  & 77.3  & 82.6 & 75.7 & 68.6 & 82.4 & 56.5 \\
    Tompson et al.~\cite{tompson14nips} & 95.8 & 90.3 & 80.5 & 74.3 & 77.6 & 69.7 & 62.8 & 79.6 & 51.8 \\
    Carreira et al.~\cite{carreira16cvpr} & 95.7  & 91.7  & 81.7  & 72.4  & 82.8  & 73.2 & 66.4 & 81.3 & 49.1\\     
    Tompson et al.~\cite{Tompson:2015:EOL}  &96.1 & 91.9 & 83.9 &77.8 & 80.9 & 72.3 & 64.8 & 82.0 & 54.9\\
    Wei et al.~\cite{wei16cvpr} & \textbf{97.8}  & 95.0  & 88.7  & 84.0  & \textbf{88.4}  & 82.8 & \textbf{79.4} & \textbf{88.5} & \textbf{61.4} \\
    \bottomrule
  \end{tabular}
  \vspace{0.1em}     
  \caption[]{Pose estimation results (PCK$_h$) on MPII Single Person.}
    \vspace{-0.5cm}
  \label{tab:multicut:mpii}
\end{table}

\myparagraph{Results on MPII Single Person.} The results are shown in
Tab.~\ref{tab:multicut:mpii}. ResNet-152 achieves $87.8$\% PCK$_h$ and
$60.0$\% AUC, while intermediate supervision slightly improves the
performance further to $88.5$\% PCK$_h$ and $60.8$\% AUC.  Comparing
the results to the state of the art we observe significant improvement
over \deepcut~\cite{pishchulin16cvpr} ($+5.9$\% PCK$_h$, $+4.2$\%
AUC), which again underlines the importance of using extremely deep
model. The proposed approach performs on par with the best know result
by Wei et al.~\cite{wei16cvpr} ($88.5$ vs. $88.5$\% PCK$_h$) for the
maximum distance threshold, while slightly loosing when using the
entire range of thresholds ($60.8$ vs. $61.4$\% AUC). We envision that
extending the proposed approach to incorporate multiple scales as
in~\cite{wei16cvpr} should improve the performance. The model trained
on the union of MPII, LSPET and LSP training sets achieves $88.3$\%
PCK$_h$ and $60.7$\% AUC. The fact that the same model achieves
similar performance on both LSP and MPII benchmarks demonstrates the
generality of our approach.

\section{Image-Conditioned Pairwise Terms}
\label{sec:pairwise}


As discussed in Sec.~\ref{sec:unary-pairwise}, a large receptive field
for the CNN-based part detectors allows to accurately predict the
presence of a body part at a given location. However, it also contains
enough evidence to reason about locations of other parts in the
vicinity. We draw on this insight and propose to also use deep
networks to make pairwise part-to-part predictions. They are
subsequently used to compute the pairwise probabilities and show
significant improvements for multi-person pose estimation.

\subsection{Model}

\begin{figure}[t]
  \centering
  \begin{tabular}{c c c c c c}
    \includegraphics[height=0.20\linewidth]{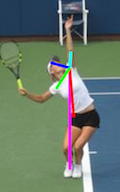}&
    \includegraphics[height=0.20\linewidth]{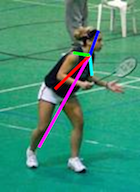}&
    \includegraphics[height=0.20\linewidth]{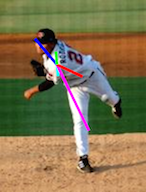}&
    \includegraphics[height=0.20\linewidth]{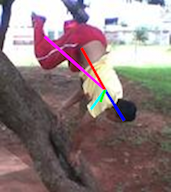}&
    \includegraphics[height=0.20\linewidth]{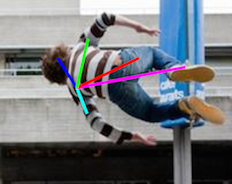}&
    \includegraphics[height=0.20\linewidth]{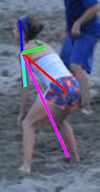} \\
    \includegraphics[height=0.20\linewidth]{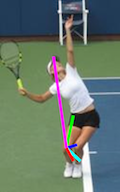}&
    \includegraphics[height=0.20\linewidth]{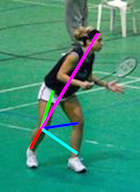}&
    \includegraphics[height=0.20\linewidth]{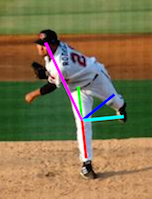}&
    \includegraphics[height=0.20\linewidth]{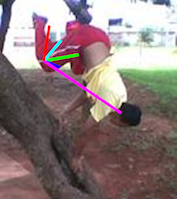}&
    \includegraphics[height=0.20\linewidth]{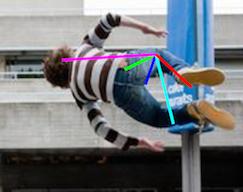}&
    \includegraphics[height=0.20\linewidth]{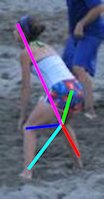} \\

  \end{tabular}
  \vspace{-1em}
  \caption{Visualizations of regression predictions. Top: from left
    shoulder to the right shoulder (green), right hip (red), left
    elbow (light blue), right ankle (purple) and top of the head (dark
    blue). Bottom: from right knee to the right hip (green), right
    ankle (red), left knee (dark blue), left ankle (light blue) and
    top of the head (purple). Longer-range predictions, such as
    e.g. shoulder -- ankle may be less accurate for harder poses (top row, images
    2 and 3) compared to the nearby predictions. However,
    they provide enough information to constrain the search space in
    the fully-connected spatial model.}
  \label{fig:regression_other_joint}
\end{figure}

\begin{figure}[t]
    \begin{center}
        \includegraphics[height=0.22\textwidth]{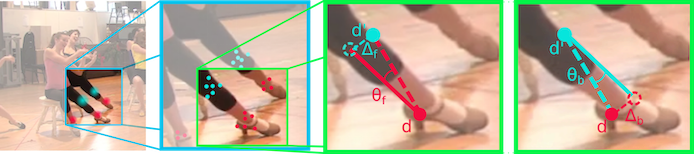}
        \caption{Visualization of features extracted to score the pairwise. See text for details.}
        \label{fig:pairwise-sketch}
        \vspace{-0.7cm}
    \end{center}
\end{figure}

Our approach is inspired by the body part location refinement described in
Sec.~\ref{sec:unary-pairwise}.  In addition to predicting
offsets for the current joint, we directly regress from the current
location to the relative positions of all other joints. For each
scoremap location $k = (x_k,y_k)$ that is marked positive w.r.t the
joint $c \in C$ and for each remaining joint $c' \in C \setminus c$, we
define a relative position of $c'$ w.r.t. $c$ as a tuple $t_{cc'}^k =
(x_{c'}-x_k, y_{c'}-x_k)$.
We add an extra layer that predicts relative position $o_{cc'}^k$ and
train it with a smooth L$_1$ loss function. We thus perform 
\emph{joint} training of body part detectors (cross-entropy loss),
location regression (L$_1$ loss) and pairwise regression (L$_1$ loss)
by linearly combining all three loss functions.
The targets $t$ are normalized to have zero mean and unit variance
over the training set. Results of such predictions are shown in
Fig.~\ref{fig:regression_other_joint}.

We then use these predictions to compute pairwise costs
$\beta_{dd'cc'}$. For any pair of detections $(d, d')$
(Fig.~\ref{fig:pairwise-sketch}) and for any pair of joints $(c,
c')$ we define the following quantities: locations $l_d$, $l_d'$ of
detections $d$ and $d'$ respectively; the offset prediction
$o_{cc'}^d$ from $c$ to $c'$ at location $d$ (solid red) coming from
the CNN and similarly the offset prediction $o_{c'c}^{d'}$ (solid
turquoise).  We then compute the offset between the two predictions:
$\hat{o}_{dd'}= l_{d'}-l_d$ (marked in dashed red).  The degree to
which the prediction $o_{cc'}^d$ agrees with the actual offset
$\hat{o}_{dd'}$ tells how likely $d$, $d'$ are of classes $c$, $c'$
respectively and belong to the same person.  We measure this by
computing the distance between the two offsets $\Delta_f = \lVert
\hat{o}_{dd'}-o_{cc'}^{d} \rVert_2$, and the absolute angle
$\theta_{f} = \lvert \measuredangle (\hat{o}_{dd'}, o_{cc'}^{d})
\rvert $ where $f$ stands for forward direction, i.e from $d$ to $d'$.
Similarly, we incorporate the prediction $o_{c'c}^{d'}$ in the backwards
direction by computing $\Delta_b = \lVert \hat{o}_{d'd}-o_{c'c}^{d'}
\rVert_2$ and $\theta_{b} = \lvert \measuredangle (\hat{o}_{d'd},
o_{c'c}^{d'}) \rvert $.  Finally, we define a feature vector by
augmenting features with exponential terms: $f_{dd'cc'} = (\Delta_f,
\theta_f, \Delta_b, \theta_b, \exp(-\Delta_f), \dots,
\exp(-\theta_b))$.

We then use the features $f_{dd'cc'}$ and define logistic model:
\begin{align}
\label{eq:z-probability}
p(z_{dd'cc'} = 1 | f_{dd'cc'}, \omega_{cc'}) & = \frac{1}{1 + \exp(-\langle \omega_{cc'}, f_{dd'cc'} \rangle)}.
\end{align}
where $K=(|C|\times (|C| + 1))/2$ parameters $\omega_{cc'}$ are estimated using ML.

\subsection{Sampling Detections}

\myparagraph{Location refinement NMS.}
\deepcut{} samples the set of detections $D$
from the scoremap by applying non-maximum suppression (NMS). Here, we
utilize location refinement and correct grid locations with the
predicted offsets before applying NMS. This pulls detections that
belong to a particular body joint towards its true location thereby
increasing the density of detections around that location, which
allows to distribute the detection candidates in a better way.

\myparagraph{Splitting of part detections.} \deepcut{} ILP solves the
clustering problem by labeling each detection $d$ with a single part
class $c$ and assigning it to a particular cluster that corresponds to
a distinct person. However, it may happen that the same spatial
location is occupied by more than one body joint, and therefore, its
corresponding detection can only be labeled with one of the respecting
classes. A naive solution is to replace a detection with $n$
detections for each part class, which would result in a prohibitive
increase in the number of detections. We simply split a detection
$d$ into several if more than one part has unary probability that is
higher than a chosen threshold $s$ (in our case $s=0.4$).

\subsection{Evaluation of Pairwise}

\myparagraph{Datasets and evaluation measure.} We evaluate on the
challenging public ``MPII Human Pose'' (``Multi-Person'')
benchmark~\cite{andriluka14cvpr} consisting of $3844$ training
and $1758$ testing groups of multiple overlapping people in highly
articulated poses with a variable number of parts. We perform all
intermediate experiments on a validation set of $200$ images sampled
uniformly at random and refer to it as MPII Multi-Person Val. We
report major results on the full testing set, and on the subset of
$288$ images for the direct comparison to~\cite{pishchulin16cvpr}. We
follow the official evaluation
protocol\footnote{http://human-pose.mpi-inf.mpg.de/\#evaluation} and
evaluate on groups using provided rough group location and scale. In
more detail, we localize each group by cropping around the group using
the provided information and use resulting crops as input to
multi-person pose estimation. The AP measure~\cite{pishchulin16cvpr}
evaluating consistent body part detections is used for performance
comparison. Additionally, we report median running time per frame
measured in seconds\footnote{Run-time is measured on a single core
  Intel Xeon $2.70$GHz}.

\tabcolsep 1.5pt
\begin{table}[tbp]
 \scriptsize
  \centering
  \begin{tabular}{@{} l l c ccc ccc c|c@{}}
    \toprule
    Unary\hspace{1.5cm} & Pairwise & Head   & Sho  & Elb & Wri & Hip & Knee & Ank & AP & time [s/frame] \\
    \midrule
    $\deepcut$~\cite{pishchulin16cvpr} & $\deepcut$~\cite{pishchulin16cvpr}
    & 50.1  & 44.1  & 33.5  & 26.5  & 33.0  & 28.5 & 14.4 & 33.3 & 259220\\ 
    $\deepcut$~\cite{pishchulin16cvpr} & this work & 68.3  & 58.3  & 47.4  & 38.9  & 45.2  & 41.8 & 31.2 & 47.7 & 1987\\
    this work & this work & \textbf{70.9}  & 59.8  &  \textbf{53.1}  &  \textbf{44.4}  & 50.0  & 46.4 & 39.5 &  52.3 & 1171\\ 
    \multicolumn{2}{l}{\quad + location refinement before NMS} & 70.3  &  \textbf{61.6}  & 52.1  & 43.7  &  \textbf{50.6}  &  \textbf{47.0} &  \textbf{40.6} & \textbf{52.6} &  \textbf{578} \\ 
    \bottomrule
  \end{tabular}
  \vspace{0.1em}
\caption[]{Effects of proposed pairwise and unaries on the pose estimation performance (AP) on MPII Multi-Person Val.}
\label{tab:multicut:mpii-multi:pairwise}
\vspace{-2.0em}
\end{table}

\tabcolsep 1.5pt
\begin{table}[tbp]
 \scriptsize
  \centering
  \begin{tabular}{@{} l c ccc ccc c|c@{}}
    \toprule
    Setting& Head   & Sho  & Elb & Wri & Hip & Knee & Ank & AP & time [s/frame] \\
    \midrule
    bi-directional + angle  &  \textbf{70.3}  &  \textbf{61.6}  &  \textbf{52.1}  & 43.7  &  \textbf{50.6}  &  \textbf{47.0} &  \textbf{40.6} &  \textbf{52.6} &  \textbf{578} \\ 
    uni-directional + angle & 69.3  & 58.4  & 51.8  &  \textbf{44.2}  & 50.4  & 44.7 & 36.3 & 51.1 & 2140 \\ 
    bi-directional   & 68.8  & 58.3  & 51.0  & 42.7  & 51.1  & 46.5 & 38.7 & 51.3 & 914 \\ 
    \bottomrule
  \end{tabular}
  \vspace{0.1em}
\caption[]{Effects of different versions of the pairwise terms on the pose estimation performance (AP) on MPII Multi-Person Val.}
\label{tab:multicut:mpii-multi:pairwise:ablation}
\vspace{-0.5cm}
\end{table}

\myparagraph{Evaluation of unaries and pairwise.} The results are
shown in Tab.~\ref{tab:multicut:mpii-multi:pairwise}. Baseline
$\deepcut$ achieves $33.3$\% AP. Using the proposed pairwise
significantly improves performance achieving $47.7$\% AP. This clearly
shows the advantages of using image-conditioned pairwise to
disambiguate the body part assignment for multiple overlapping
individuals. Remarkably, the proposed pairwise dramatically reduce the
run-time by two orders of magnitude ($1987$ vs. $259220$
s/frame). This underlines the argument that using strong pairwise in
the fully-connected model allows to significantly speed-up the
inference. Using additionally the proposed part detectors further
boosts the performance ($52.3$ vs. $47.7$\% AP), which can be
attributed to better quality part hypotheses. Run-time is again almost
halved, which clearly shows the importance of obtaining high-quality
part detection candidates for more accurate and faster inference.
Performing location refinement before NMS slightly improves the
performance, but also reduces the run-time by $2$x: this allows to
increase the density of detections at the most probable body part
locations and thus suppresses more detections around the most confident
ones, which leads to better distribution of part detection candidates
and reduces confusion generated by the near-by detections. Overall, we
observe significant performance improvement and dramatic reduction in
run-time by the proposed $\deepercut$ compared to the baseline
$\deepcut$.

\myparagraph{Ablation study of pairwise.} An ablation study of the
proposed image-conditioned pairwise is performed in
Tab.~\ref{tab:multicut:mpii-multi:pairwise:ablation}. Regressing from
both joints onto the opposite joint's location and including angles
achieves the best performance of $52.6$\% AP and the minimum run-time
of $578$ s/frame. Regressing from a single joint only slightly reduces the
performance to $51.1$\% AP, but significantly increases run-time by
$4$x: these pairwise are less robust compared to the bi-directional,
which confuses the inference. Removing the angles from the pairwise
features also decreases the performance ($51.3$ vs. $52.6$\% AP) and
doubles run-time, as it removes the information about body part
orientation.

\vspace{-0.25cm}
\section{Incremental Optimization}
\vspace{-0.15cm}
\label{sec:incremental}
Solving one instance of the \deepcut{} ILP for all body part candidates detected for an image, 
as suggested in~\cite{pishchulin16cvpr} and summarized in Sec.~\ref{sec:deepcut},
is elegant in theory but disadvantageous in practice:

Firstly, the time it takes to compute constant-factor approximative feasible solution
by the branch-and-cut algorithm~\cite{pishchulin16cvpr}
can be exponential in the number of body part candidates in the worst case.
%
In practice, this limits the number of candidates that can be processed by this algorithm.
Due to this limitation, it does happen that body parts and, for images showing many persons, entire persons 
are missed, simply because they are not contained in the set of candidates.

Secondly, solving one instance of the optimization problem for the entire image means that
no distinction is made between part classes detected reliably, e.g.~head and shoulders,
and part classes detected less reliably, e.g.~wrists, elbows and ankles.
Therefore, it happens that unreliable detections corrupt the solution.

To address both problems, we solve not one instance of the \deepcut{} ILP
but several, starting with only those body part classes that are detected most reliably
and only then considering body part classes that are detected less reliably.
Concretely, we study two variants of this incremental optimization approach which are defined in
Tab.~\ref{table:incremental-optimization}.
Specifically, the procedure works as follows:

For each subset of body part classes defined in 
Tab.~\ref{table:incremental-optimization}, 
an instance of the \deepcut{} ILP is set up
and a constant-factor approximative feasible solution computed using the branch-and-cut algorithm. 
This feasible solution selects, labels and clusters a subset of part candidates, namely of those part classes that are considered in this instance.
For the next instance,
each cluster of body part candidates of the same class from the previous instance becomes just one part candidate whose class is fixed.
Thus, the next instance is an optimization problem for selecting, labeling and clustering 
body parts that have not been determined by previous instances. 
Overall, this allows us to start with more part candidates consistently and thus improve the pose estimation result significantly.
%

\begin{table}[t]
\centering
\scriptsize
\begin{tabular}{l@{\hspace{4ex}} l@{\hspace{4ex}} l@{\hspace{4ex}}l}
\toprule
& Stage 1& Stage 2 & Stage 3 \\
\midrule
2-stage 	

    & head, shoulders
	& hips, knees
	& \\

    & elbows, wrists
    & ankles
    & \\
    
\midrule
3-stage 

& head
& elbows
& hips, knees \\

& shoulders
& wrists
& ankles\\

\bottomrule

\end{tabular}
    \vspace{0.1em}
\caption{As the run-time of the DeepCut branch-and-cut algorithm
  limits the number of part candidates that can be processed in
  practice, we split the set of part classes into subsets, coarsely
  and finely, and solve the pose estimation problem incrementally.}
    \vspace{-1.5em}
\label{table:incremental-optimization}

\end{table}

\subsection{Evaluation of Incremental Optimization}

\tabcolsep 1.5pt
\begin{table}[tbp]
 \scriptsize
  \centering
  \begin{tabular}{@{} l c ccc ccc c|c@{}}
    \toprule
    Setting& Head   & Sho  & Elb & Wri & Hip & Knee & Ank & AP & time [s/frame]\\
    \midrule
    1-stage optimize, 100 det, nms 1x & 70.3  & 61.6  & 52.1  & 43.7  & 50.6  & 47.0 & 40.6 & 52.6 & 578 \\ 
    1-stage optimize, 100 det, nms 2x & 71.3  & 64.1  & 55.8  & 44.1  & 53.8  & 48.7 & 41.3 & 54.5 & 596 \\ 
    1-stage optimize, 150 det, nms 2x & 74.1  & 65.6  & 56.0  & 44.3  & 54.4  & 49.2 & 39.8 & 55.1 & 1041 \\ 
    \midrule
    2-stage optimize & 75.9  & 66.8  & 58.8  & 46.1  & 54.1  & 48.7 & 42.4 & 56.5 & 483 \\ 
    3-stage optimize & 78.3  & 69.3  & 58.4  & 47.5  & 55.1  & 49.6 & 42.5 & 57.6 & 271 \\ 
    \quad + split detections &  \textbf{78.5}  &  \textbf{70.5}  &  \textbf{59.7}  &  \textbf{48.7}  &  \textbf{55.4}  & \textbf{50.6} & \textbf{44.4} & \textbf{58.7} & \textbf{270} \\ 
    
    \midrule
    $\deepcut$~\cite{pishchulin16cvpr}
& 50.1  & 44.1  & 33.5  & 26.5  & 33.0  & 28.5 & 14.4 & 33.3 & 259220\\ 
    \bottomrule
  \end{tabular}
  \vspace{0.1em}
\caption[]{Performance (AP) of different  hierarchical versions of $\deepercut$ on MPII Multi-Person Val.}
\label{tab:multicut:mpii-multi:hierarchical}
\vspace{-0.65cm}
\end{table}

Results are shown in
Tab.~\ref{tab:multicut:mpii-multi:hierarchical}. Single stage
optimization with $|D|=100$ part detection candidates achieves
$52.6$\% AP (best from
Tab.~\ref{tab:multicut:mpii-multi:pairwise}). More aggressive NMS with
radius of $24$ px improves the performance ($54.5$ vs. $52.6$\% AP),
as it allows to better distribute detection candidates. Increasing
$|D|$ to $150$ slightly improves the performance by +0.6\% AP, but
significantly increases run-time ($1041$ vs. $596$ s/frame). We found
$|D|=150$ to be maximum total number of detection candidates ($11$ per
part) for which optimization runs in a reasonable time. Incremental
optimization of 2-stage inference slightly improves the performance
($56.5$ vs. $55.1$\% AP) as it allows for a larger number of detection
candidates per body part ($20$) and leverages typically more confident
predictions of the upper body parts in the first stage before solving
for the entire body. Most importantly, it halves the median run-time
from $1041$ to $483$ s/frame. Incremental optimization of 3-stage
inference again almost halves the run-time to $271$ s/frame while noticeably
improving the human pose estimation performance for all body parts but
elbows achieving 57.6\% AP.
These results clearly demonstrate the advantages of the proposed
incremental optimization. Splitting the detection candidates that
simultaneously belong to multiple body parts with high confidence
slightly improves the performance to $58.7$\% AP. This helps to
overcome the limitation that each detection candidate can be assigned
to a single body part and improves on cases where two body parts
overlap thus sharing the same detection candidate.  We also compare
the obtained results to \deepcut{} in
Tab.~\ref{tab:multicut:mpii-multi:hierarchical} (last row). The
proposed \deepercut{} outperforms baseline \deepcut{} (58.7 vs. 33.3\% AP)
by almost doubling the performance, while run-time is reduced
dramatically by $3$ orders of magnitude from the infeasible $259220$ s/frame
to affordable $270$ s/frame. This comparison clearly demonstrates the power
of the proposed approach and dramatic effects of better unary,
pairwise and optimization on the overall pose estimation performance
and run-time.
\vspace{-0.25cm}
\subsection{Comparison to the State of the Art}
We compare to others on MPII Multi-Person Test and
WAF~\cite{eichner10eccv} datasets.

\myparagraph{Results on MPII Multi-Person.} For direct comparison
with~\deepcut{} we evaluate on the same subset of $288$ testing images
as in~\cite{pishchulin16cvpr}. Additionally, we provide the results on
the entire testing set. Results are shown in
Tab.~\ref{tab:multicut:mpii-multi}. \deepercut{} without incremental
optimization already outperforms \deepcut{} by a large margin ($66.2$
vs. $54.1$\% AP). Using 3-stage incremental optimization further
improves the performance to $69.7$\% AP improving by a dramatic
$16.5$\% AP over the baseline. Remarkably, the run-time is reduced
from $57995$ to $230$ s/frame, which is an improvement by two orders
of magnitude. Both results underline the importance of strong
image-conditioned pairwise terms and incremental optimization to
maximize multi-person pose estimation performance at the reduced
run-time. A similar trend is observed on the full set: 3-stage
optimization improves over a single stage optimization ($59.4$
vs. $54.7$\% AP). We observe that the performance on the entire
testing set is over $10$\% AP lower compared to the subset and
run-time is doubled. This implies that the subset of $288$ images is
easier compared to the full testing set. We envision that performance
differences between \deepercut{} and \deepcut{} on the entire set will
be at least as large as when compared on the subset. We also compare
to a strong two-stage baseline: first each person is pre-localized by
applying the state-of-the-art detector~\cite{ren2015faster} following
by NMS and retaining rectangles with scores at least $0.8$; then pose
estimation for each rectangle is performed using \deepercut{} unary
only. Being significantly faster ($1$ s/frame) this approach reaches
$51.0$\% AP vs. $59.4$\% AP by \deepercut{}, which clearly shows the
power of joint reasoning by the proposed approach.

\myparagraph{Excluding out-of-group predictions.}
Qualitative analysis of group crops generated by our cropping
procedure reveals that crops often include people from other groups
due to excessive crop sizes. \deepercut{} is being penalized for
correctly predicting poses of such individuals outside of each group:
in AP-based evaluation such predictions are treated as false positives
thus significanly decreasing the AP performance. In order to address
this issue we exclude such predictions in the following way: we extend
provided bounding box aroud each group's center by a constant padding
($37$ pixels in the scale-normalized image), and use the extended
bounding box to filter out all predicted poses whose centers of mass
fall outside of the bounding box. This significantly
improves \deepercut{} results ($70.0$ vs. $59.4$\% AP), as well as the
results of the baseline two-stage approach ($59.7$ vs. $51.0$\% AP).
Proposed \deepercut{} significantly improves over the strong two-stage
baseline ($70.0$ vs. $59.7$\% AP), which is in agreement with our
observations reported above.

\tabcolsep 1.5pt
\begin{table}[tbp]
 \scriptsize
  \centering
  \begin{tabular}{@{} l c ccc ccc c|c@{}}
    \toprule
    Setting& Head   & Sho  & Elb & Wri & Hip & Knee & Ank & AP & time [s/frame]\\
    \midrule
    \multicolumn{10}{c}{subset of 288 images as in~\cite{pishchulin16cvpr}}\\
    $\deepercut$ (1-stage) & 83.3  & 79.4  & 66.1  & 57.9  & 63.5  & 60.5 & 49.9 & 66.2 & 1333 \\ 
    $\deepercut$ & \textbf{87.5}  & \textbf{82.8}  & \textbf{70.2}  & \textbf{61.6}  & \textbf{66.0}  & \textbf{60.6} & \textbf{56.5} & \textbf{69.7} & \textbf{230} \\ 
    $\deepcut$~\cite{pishchulin16cvpr} & 73.4  & 71.8  & 57.9  & 39.9  & 56.7  & 44.0 & 32.0 & 54.1 & 57995\\ 
    \midrule
    \multicolumn{10}{c}{full set}\\
    $\deepercut$ (1-stage) & 73.7  & 65.4  & 54.9  & 45.2  & 52.3  & 47.8 & 40.7 & 54.7 & 2785 \\ 
    $\deepercut$                & \textbf{79.1}  & \textbf{72.2}  & \textbf{59.7}  & \textbf{50.0}  & \textbf{56.0}  & \textbf{51.0} & \textbf{44.6} & \textbf{59.4} & 485 \\ 
    Faster R-CNN~\cite{ren2015faster} + unary &64.9  & 62.9  & 53.4  &
           44.1  & 50.7  & 43.1 & 35.2 & 51.0 & \textbf{1}\\ 

    \midrule
    \multicolumn{10}{c}{full set, excluding out-of-group predictions}\\
    $\deepercut$                & \textbf{89.4}  & \textbf{84.5}  & \textbf{70.4}  & \textbf{59.3}  & \textbf{68.9}  & \textbf{62.7} & \textbf{54.6} & \textbf{70.0} & 485 \\ 
    Faster R-CNN~\cite{ren2015faster} + unary &75.1  & 73.6  & 62.7  & 51.0  & 61.1  & 52.6 & 42.2 & 59.7 & \textbf{1}\\ 

    \bottomrule
  \end{tabular}
  \vspace{0.1em}
\caption[]{Pose estimation results (AP) on MPII Multi-Person.}
\label{tab:multicut:mpii-multi}
\vspace{-0.5cm}
\end{table}

\tabcolsep 1.5pt
\begin{table}[tbp]
 \scriptsize
  \centering
  \begin{tabular}{@{} l c cc cc |c@{}}
    \toprule
    Setting& Head   & U Arms  & L Arms & Torso & $m$PCP  & AOP \\
    \midrule
    $\deepercut$ nms 3.0 & \textbf{99.3} & \textbf{83.8} & \textbf{81.9} & \textbf{87.1} & \textbf{86.3} & \textbf{88.1} \\
    \midrule
    \deepcut~\cite{pishchulin16cvpr} & \textbf{99.3} & 81.5 & 79.5 & \textbf{87.1} & 84.7 & 86.5 \\
    Ghiasi et al.~\cite{ghiasi14cvpr}  & -      & -      & -      & -         & 63.6  &  74.0 \\
    Eichner\&Ferrari~\cite{eichner10eccv}   & 97.6   & 68.2  &   48.1     &86.1    &69.4   & 80.0 \\
    Chen\&Yuille~\cite{Chen:2015:POC}  & 98.5 &   77.2 &  71.3 &   88.5 &   80.7 &   84.9 \\
    \bottomrule
  \end{tabular}
    \vspace{0.1em}
  \caption[]{Pose estimation results ($m$PCP) on WAF dataset.}
    \vspace{-1.0em}
  \label{tab:multicut:waf}
\end{table}

\myparagraph{Results on WAF.} Results using the official evaluation
protocol~\cite{eichner10eccv} assuming $m$PCP and AOP evaluation
measures and considering detection bounding boxes provided
by~\cite{eichner10eccv} are shown in
Tab.~\ref{tab:multicut:waf}. \deepercut{} achieves the best result
improving over the state of the art \deepcut{} ($86.3$ vs. $84.7$\%
$m$PCP, $88.1$ vs. $86.5$\% AOP). Noticeable improvements are observed
both for upper (+$2.3$\% $m$PCP) and lower ($+2.4$\% $m$PCP)
arms. However, overall performance differences between $\deepercut$
and the baseline $\deepcut$ are not as pronounced compared to MPII
Multi-Person dataset. This is due to the fact that actual differences
are washed out by the peculiarities of the $m$PCP evaluation measure:
$m$PCP assumes that people are pre-detected and human pose estimation
performance is evaluated only for people whose upper body detections
match the ground truth. Thus, a pose estimation method is not
penalized for generating multiple body pose predictions, since the
only pose prediction is considered whose upper body bounding box best
matches the ground truth. We thus re-evaluate the competing
approaches~\cite{pishchulin16cvpr,Chen:2015:POC} using the more
realistic AP evaluation measure\footnote{We used publicly-available
pose predictions of~\cite{Chen:2015:POC} for all people in WAF
dataset.}. The results are shown in
Tab.~\ref{tab:multicut:waf:ap}. $\deepercut$ significantly improves
over $\deepcut$ ($82.0$ vs. $76.2$\% AP). The largest boost in
performance is achieved for head ($+16.0$\% AP) and wrists ($+5.2$\%
AP): $\deepercut$ follows incremental optimization strategy by first
solving for the most reliable body parts, such as head and shoulders,
and then using the obtained solution to improve estimation of harder
body parts, such as wrists. Most notably, run-time is dramatically
reduced by $3$ orders of magnitude from $22000$ to $13$ s/frame. These
results clearly show the advantages of the proposed approach when
evaluated in the real-world detection setting. The
proposed \deepercut{} also outperforms
\cite{Chen:2015:POC} by a large margin. The performance difference is
much more pronounced compared to using $m$PCP evaluation measure: in
contrast to $m$PCP, AP penalizes multiple body pose predictions of the
same person. We envision that better NMS strategies are likely to
improve the AP performance of \cite{Chen:2015:POC}.

\tabcolsep 1.5pt
\begin{table}[tbp]
 \scriptsize
  \centering
  \begin{tabular}{@{} l c ccc c|c@{}}
    \toprule
    Setting& Head   & Sho  & Elb & Wri & AP & time [s/frame]\\
    \midrule
    $\deepercut$  & \textbf{92.6}  & \textbf{81.1}  & \textbf{75.7}  & \textbf{78.8} & \textbf{82.0} & \textbf{13}\\
    $\deepcut$~\cite{pishchulin16cvpr}& 76.6  & 80.8  & 73.7  & 73.6  & 76.2 & 22000\\
    Chen\&Yuille~\cite{Chen:2015:POC}  & 83.3  & 56.1 & 46.3  & 35.5 & 55.3 & - \\
    \bottomrule
  \end{tabular}
  \vspace{0.1em}     
  \caption[]{Pose estimation results (AP) on WAF dataset.}
    \vspace{-2.5em}
  \label{tab:multicut:waf:ap}
\end{table}

\vspace{-0.25cm}
\section{Conclusion}
\vspace{-0.15cm}
\label{sec:conclusion}
In this paper we significantly advanced the state of the art in
articulated multi-person human pose estimation. To that end we
carefully re-designed and thoroughly evaluated several key
ingredients. First, drawing on the recent advances in deep learning we
proposed strong extremely deep body part detectors that -- taken alone
-- already allow to obtain state of the art performance on standard
pose estimation benchmarks. Second, we introduce novel
image-conditioned pairwise terms between body parts that allow to
significantly push the performance in the challenging case of
multi-people pose estimation, and dramatically reduce the run-time of
the inference in the fully-connected spatial model. Third, we
introduced a novel incremental optimization strategy to further reduce
the run-time and improve human pose estimation accuracy. Overall, the
proposed improvements allowed to almost double the pose estimation
accuracy in the challenging multi-person case while reducing the
run-time by $3$ orders of magnitude.

\begin{subappendices}
\renewcommand{\thesection}{\Alph{section}}
\section{Qualitative Evaluation on MPII Multi-Person Dataset}
We perform qualitative analysis of the proposed \deepercut{} approach
on MPII Multi-Person dataset. First, we visualize novel
image-conditioned pairwise terms. Then, we demonstrate successful and
failure cases of the proposed approach.

\subsection{Image-Conditioned Pairwise}
\begin{figure*}
  \centering

  \begin{tabular}{c c c c c}
    &&
    \includegraphics[height=0.17\linewidth]{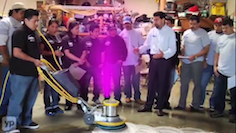} &
    &
    \includegraphics[height=0.17\linewidth]{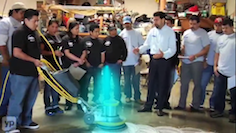}
    \\[-3pt]
    &&
    $p(\textrm{r knee} \; | \; \textrm{r ankle})$ &
    &
    $p(\textrm{r knee} \; | \; \textrm{r hip})$
    \\[5pt]
    &&
    \includegraphics[height=0.17\linewidth]{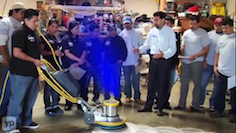} &
    \includegraphics[height=0.17\linewidth]{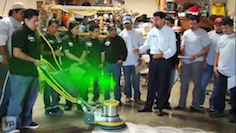} &
    \includegraphics[height=0.17\linewidth]{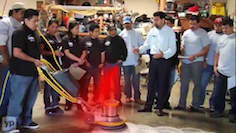} 
    \\[-3pt]
    &&
    $p(\textrm{r knee} \; | \; \textrm{l ankle})$ &
    $p(\textrm{r knee} \; | \; \textrm{l knee})$ &
    $p(\textrm{r knee} \; | \; \textrm{l hip})$
    \\[5pt]
    \begin{sideways}\bf \quad pairwise\end{sideways}&
    \begin{sideways}\bf \quad per part\end{sideways}&
    \includegraphics[height=0.170\linewidth]{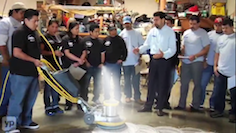} &
    \includegraphics[height=0.170\linewidth]{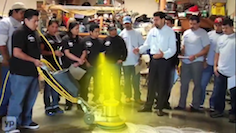} &
    \includegraphics[height=0.170\linewidth]{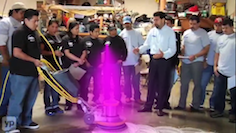}
    \\[-3pt]
    &&
    $p(\textrm{r knee} \; | \; \textrm{r wrist})$ &
    $p(\textrm{r knee} \; | \; \textrm{r elbow})$ &
    $p(\textrm{r knee} \; | \; \textrm{r shoulder})$
    \\[5pt]
    &&
    \includegraphics[height=0.170\linewidth]{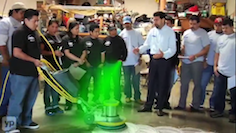} &
    \includegraphics[height=0.170\linewidth]{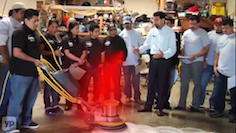} &
    \includegraphics[height=0.170\linewidth]{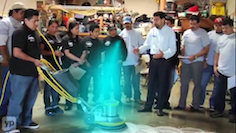}
    \\[-3pt]
    &&
    $p(\textrm{r knee} \; | \; \textrm{l wrist})$ &
    $p(\textrm{r knee} \; | \; \textrm{l elbow})$ &
    $p(\textrm{r knee} \; | \; \textrm{l shoulder})$
    \\[5pt]
    &&
    \includegraphics[height=0.170\linewidth]{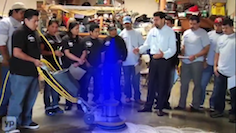} &
    \includegraphics[height=0.170\linewidth]{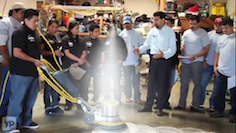} &
    \\[-3pt]
    &&
	$p(\textrm{r knee} \; | \; \textrm{chin})$ &
	$p(\textrm{r knee} \; | \; \textrm{top head})$ &
    \\[5pt]
    \midrule
    \begin{sideways}\bf \quad pairwise\end{sideways}&
    \begin{sideways}\bf \quad combined\end{sideways}&
    \multicolumn{3}{c}{\includegraphics[height=0.170\linewidth]{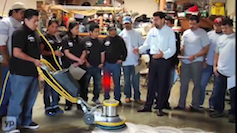}}\\
    \midrule
    \begin{sideways}\bf \quad unaries\end{sideways}&
    \begin{sideways}\bf \quad \end{sideways}&
    \multicolumn{3}{c}{\includegraphics[height=0.170\linewidth]{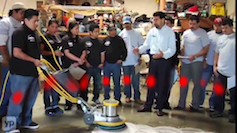}}\\
  \end{tabular}
  \vspace{-1em}
  \caption{Visualizations of the pairwise probabilities when the right
    knee of the middle person is used as target and fixed locations of
    other body parts as source. See text for explanation.}
  \label{fig:qualitative_mpii_pairwise}
\end{figure*}

\begin{figure*}
  \centering

  \begin{tabular}{c c c c c}
    \begin{sideways}\bf \quad pairwise\end{sideways}&
    \begin{sideways}\bf \quad combined\end{sideways}&
    \includegraphics[height=0.175\linewidth]{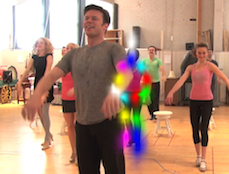} &
    \includegraphics[height=0.175\linewidth]{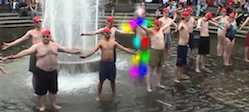} &
    \includegraphics[height=0.175\linewidth]{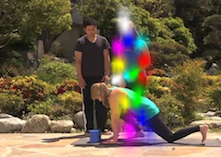}  
    \\
    \begin{sideways}\bf \quad unaries\end{sideways}&
    \begin{sideways}\bf \quad \end{sideways}&
    \includegraphics[height=0.175\linewidth]{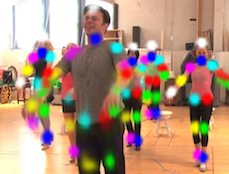}&
    \includegraphics[height=0.175\linewidth]{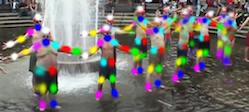}&
    \includegraphics[height=0.175\linewidth]{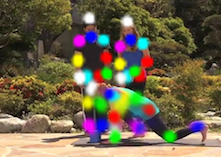}
    \\[10pt]
  \end{tabular}

  \begin{tabular}{c c c c c c}
    \begin{sideways}\bf \quad pairwise\end{sideways}&
    \begin{sideways}\bf \quad combined\end{sideways}&
    \includegraphics[height=0.180\linewidth]{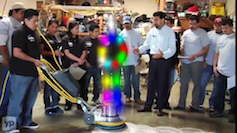} &
    \includegraphics[height=0.180\linewidth]{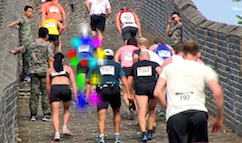} &
    \includegraphics[height=0.180\linewidth]{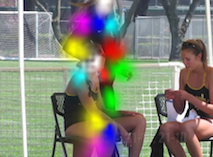}
    \\
    \begin{sideways}\bf \quad unaries\end{sideways}&
    \begin{sideways}\bf \quad \end{sideways}&
    \includegraphics[height=0.180\linewidth]{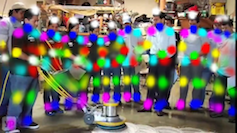} &
    \includegraphics[height=0.180\linewidth]{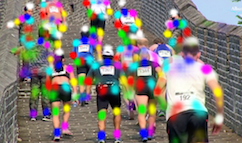} &
    \includegraphics[height=0.180\linewidth]{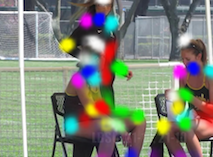} 
    \\[10pt]
  \end{tabular}
  
  \begin{tabular}{c c c c c c}
    \begin{sideways}\bf \quad pairwise\end{sideways}&
    \begin{sideways}\bf \quad combined\end{sideways}&
    \includegraphics[height=0.180\linewidth]{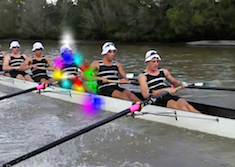} &
    \includegraphics[height=0.180\linewidth]{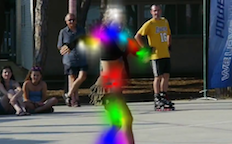} &
    \includegraphics[height=0.180\linewidth]{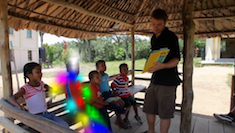} 
    \\
    \begin{sideways}\bf \quad unaries\end{sideways}&
    \begin{sideways}\bf \quad \end{sideways}&
    \includegraphics[height=0.180\linewidth]{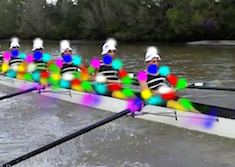} &
    \includegraphics[height=0.180\linewidth]{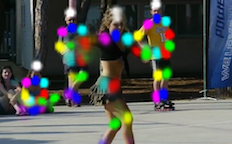} &
    \includegraphics[height=0.180\linewidth]{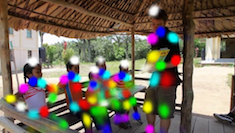} 
    \\
  \end{tabular}
  \vspace{-1em}
  \caption{Visualizations of all unary scoremaps (rows 2, 4 and 6) and
    all combined pairwise scoremaps (rows 1, 3 and 5). Pairwise
    scoremaps are visualized for a randomly selected individual to
    avoid clutter: for each target body part individual predictions
    from all other body parts are obtained and combined the same way
    as in Fig.~\ref{fig:qualitative_mpii_pairwise}. Color-coding
    corresponds to different body parts. Multiple scoremaps are
    overlaid on the same image for visualization purposes.}
  \label{fig:qualitative_mpii_pairwise_all}
\end{figure*}

In order to visually analyze the proposed pairwise terms for a
particular body part $c$, we proceed as follows. First, we randomly
select a person in the image. Next, for each body part $c'\neq c$ we
fix location of $c'$ at its ground truth location (to separate the
effects of possible misdetection) and predict the location of $c$
using the learned regressor. Then, we compute the pairwise probability
$p(cc')$\footnote{We intentionally simplify the notation compared
to Eq.~\eqref{eq:z-probability}.} for every possible location of $c$ in the
image. Fig.~\ref{fig:qualitative_mpii_pairwise}, rows $1$--$5$, shows
the probability of $c=$right knee anywhere in the image given the
fixed location of other body parts. It can be seen that individual
pairwise scoremaps have shape of cone extending towards the correct
location, but are visually quite fuzzy. In order to visualize the
effects of the pairwise predictions interacting in the fully-connected
spatial model, we combine individual scoremaps by multiplying
them. Combined scoremap is shown in
Fig.~\ref{fig:qualitative_mpii_pairwise}, row $6$. Clearly, interplay
of individual pairwise scoremaps produces a strong evidence for a
\textit{single} right knee of a selected individual. This is in
contrast to multi-modal unary scoremaps
(Fig.~\ref{fig:qualitative_mpii_pairwise}, row $7$) that show a strong
response of \textit{any} knee in the image. This clearly shows that
the proposed pairwise terms can be successfully used to filter out
body part locations belonging to multiple people thus effectively
disambiguating between
individuals. Fig.~\ref{fig:qualitative_mpii_pairwise_all} shows more
examples of combined pairwise scoremaps for each body part of randomly
selected individuals, and unary scoremaps. It can be seen that
although combined pairwise scoremaps are more fuzzy compared to the
unary scoremaps, they allow to filter out the body part detections of
other individuals when predicting the pose of the person in question.

\subsection{Examples of Successful and Failure Cases}

\begin{figure*}
  \centering
  \begin{tabular}{c c c c}
    \includegraphics[height=0.153\linewidth]{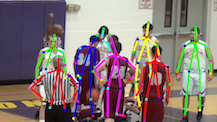} &
    \includegraphics[height=0.153\linewidth]{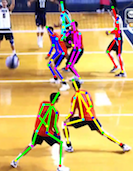} &
    \includegraphics[height=0.153\linewidth]{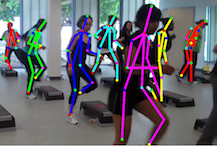} &
    \includegraphics[height=0.153\linewidth]{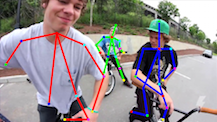} \\
    1&2&3&4\\
  \end{tabular}

  \begin{tabular}{c c c c}
    \includegraphics[height=0.200\linewidth]{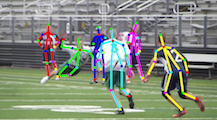} &
    \includegraphics[height=0.200\linewidth]{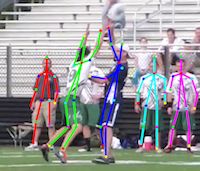} &
    \includegraphics[height=0.200\linewidth]{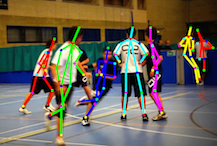} \\
    5&6&7\\
  \end{tabular}
  
  \begin{tabular}{c c c c}
    \includegraphics[height=0.16\linewidth]{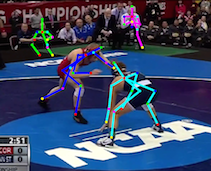} &
    \includegraphics[height=0.16\linewidth]{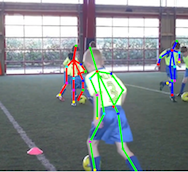} &
    \includegraphics[height=0.16\linewidth]{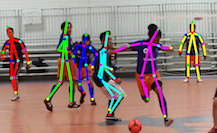} &
    \includegraphics[height=0.16\linewidth]{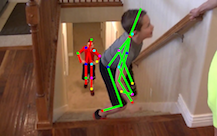} \\
    8&9&10&11\\
  \end{tabular}

  \begin{tabular}{c c c c}
    \includegraphics[height=0.163\linewidth]{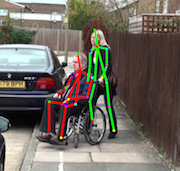} &
    \includegraphics[height=0.163\linewidth]{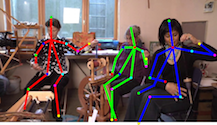} &
    \includegraphics[height=0.163\linewidth]{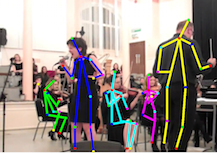} &
    \includegraphics[height=0.163\linewidth]{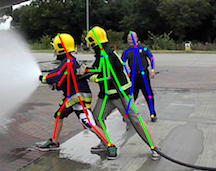} \\
    12&13&14&15\\
  \end{tabular}

  \begin{tabular}{c c c}
    \includegraphics[height=0.186\linewidth]{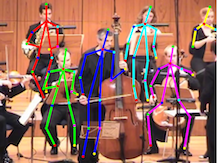} &
    \includegraphics[height=0.186\linewidth]{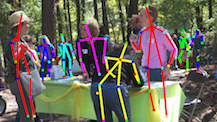} &
    \includegraphics[height=0.186\linewidth]{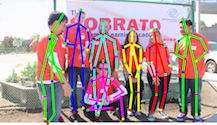} \\
    16&17&18\\
  \end{tabular}

  \begin{tabular}{c c c c}
    \includegraphics[height=0.156\linewidth]{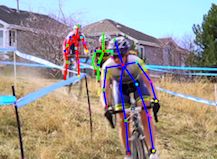} &
    \includegraphics[height=0.156\linewidth]{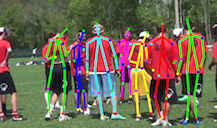} &
    \includegraphics[height=0.156\linewidth]{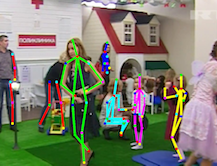} &
    \includegraphics[height=0.156\linewidth]{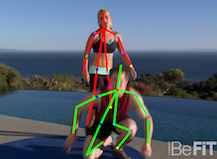} \\
    19&20&21&22\\
  \end{tabular}

  \begin{tabular}{c c c c}
    \includegraphics[height=0.154\linewidth]{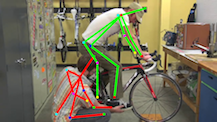} &
    \includegraphics[height=0.154\linewidth]{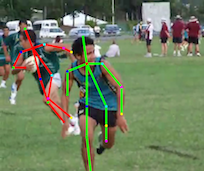} &
    \includegraphics[height=0.154\linewidth]{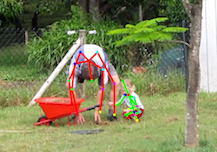} &
    \includegraphics[height=0.154\linewidth]{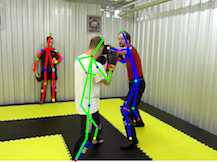} \\
    23&24&25&26\\
  \end{tabular}
  \vspace{-1em}
  \caption{Examples of successful pose estimation results obtained with our \deepercut{} model on
    the MPII Multi-Person dataset.}
  \label{fig:qualitative_mpii_good}
\end{figure*}

In Fig.~\ref{fig:qualitative_mpii_good} we include additional examples
of successful pose estimation results by \deepercut{}. The proposed
approach correctly resolves cases with only subset of body parts of a
person visible in the image. For example, in the image $4$ it correctly
outputs only visible parts for the person shown in red. \deepercut{}
is also able to correctly assemble body parts even for rare body
articulations as in the image $25$. Remarkably, the proposed approach
also correctly handles cases of strong partial occlusions. For
example, in the image $1$ body parts are correctly associated to
subjects shown in blue, cyan and magenta, even though only small
portion of the cyan subject is visible in the image.

\begin{figure*}
  \centering
  \begin{tabular}{c c c c c c}
  \begin{sideways}\bf people\end{sideways}&
    \begin{sideways}\bf \quad \end{sideways}&
      \includegraphics[height=0.170\linewidth]{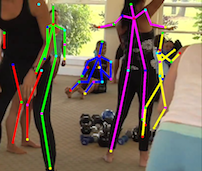} &
      \includegraphics[height=0.170\linewidth]{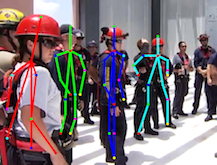} &
      \includegraphics[height=0.170\linewidth]{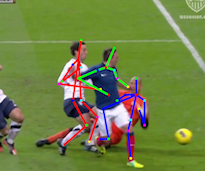} &
      \includegraphics[height=0.170\linewidth]{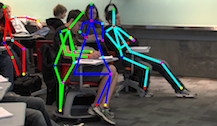} \\
      &&1&2&3&4\\
  \end{tabular}
  \begin{tabular}{c c c c c c}
    \begin{sideways}\bf \quad limbs across \end{sideways}&
    \begin{sideways}\bf \quad \end{sideways}&
      \includegraphics[height=0.200\linewidth]{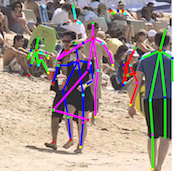} &
      \includegraphics[height=0.200\linewidth]{imgidx_0650_sticks} &
      \includegraphics[height=0.200\linewidth]{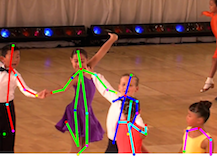} &
      \includegraphics[height=0.200\linewidth]{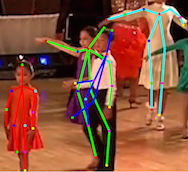} \\
      &&5&6&7&8\\
      \midrule   
  \end{tabular}
  \begin{tabular}{c c c c c c c}
    \begin{sideways}\bf symmetry\end{sideways}&
    \begin{sideways}\bf confusion\end{sideways}&
      \includegraphics[height=0.150\linewidth]{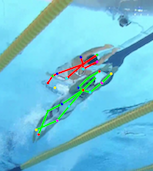} &
      \includegraphics[height=0.150\linewidth]{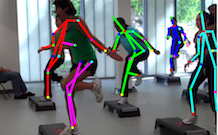} &
      \includegraphics[height=0.150\linewidth]{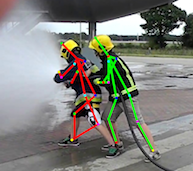} &
      \includegraphics[height=0.150\linewidth]{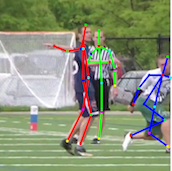} &
      \includegraphics[height=0.150\linewidth]{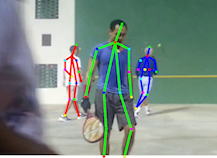} \\
      &&9&10&11&12&13\\
      \midrule   
  \end{tabular}
  \begin{tabular}{c c c c c}
    \begin{sideways}\bf \quad hard poses \end{sideways}&
    \begin{sideways}\bf \quad \end{sideways}&
      \includegraphics[height=0.190\linewidth]{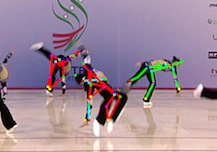} &
      \includegraphics[height=0.190\linewidth]{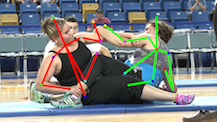} &
      \includegraphics[height=0.190\linewidth]{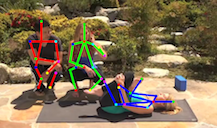} \\
      &&14&15&16\\
    \midrule 
  \end{tabular}
    \begin{tabular}{c c c c c c}
    \begin{sideways}\bf hallucinated \end{sideways}&
    \begin{sideways}\bf \quad clusters\end{sideways}&
    \includegraphics[height=0.147\linewidth]{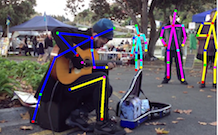} &
    \includegraphics[height=0.147\linewidth]{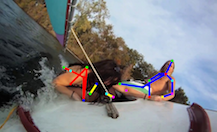} &
    \includegraphics[height=0.147\linewidth]{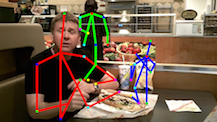} &
    \includegraphics[height=0.147\linewidth]{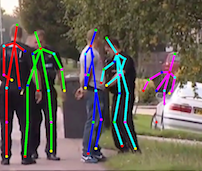} \\
    &&17&18&19&20\\
  \end{tabular}
  \vspace{-1em}
  \caption{Failure cases of our \deepercut{} model on the MPII Multi-Person dataset.}
  \label{fig:qualitative_mpii_bad}
\end{figure*}

We illustrate and analyze the failure cases of \deepercut{} in
Fig.~\ref{fig:qualitative_mpii_bad}. The included examples further
illustrate the difficulty of the task of jointly estimating body
articulations of multiple people. We identify several prominent
failure modes, and include examples for each mode. First row shows
examples of cases when \deepercut{} generates a body configuration by
merging body parts of several people. In these examples the proposed
pairwise terms failed to disambiguate between people due to their
close proximity in the image. For example, in the image $7$
\deepercut{} groups left and right limbs of the dancing pair. In the
resulting configuration the positions of upper limbs are geometrically
consistent with each other, but are not consistent with respect to
their appearance. Modeling consistency in appearance between left and
right extremities should help to mitigate this type of errors and we
will aim to address this in the future work. Another type of errors
shown at the bottom of Fig.~\ref{fig:qualitative_mpii_bad} corresponds
to cases when \deepercut{} outputs spurious body configurations that
can not be assigned to any ground-truth annotation. This happens
either when body parts of the same person are grouped into several
distinct clusters, or when a consistent body configuration is formed
from detections in background. Both of these cases are penalized by
the AP evaluation measure. In Fig.~\ref{fig:qualitative_mpii_bad} we
also visualize examples of failures due to confusion between similarly
looking left and right limbs as in the image $10$ and occasional failures
of pose estimation on rare body configurations as in image $14$.

\end{subappendices}
\bibliographystyle{splncs}
\bibliography{biblio}
\end{document}